\documentclass[11pt, letterpaper, logo, copyright]{googledeepmind}

\usepackage[utf8]{inputenc} 
\usepackage[T1]{fontenc}    
\usepackage{xcolor}
\usepackage{listings}
\usepackage{makecell}
\usepackage{hyperref}       
\usepackage{url}            
\usepackage{booktabs}       
\usepackage{amsfonts}       
\usepackage{nicefrac}       
\usepackage{microtype}      
\usepackage{xcolor}         
\usepackage{enumitem}
\usepackage{booktabs}
\usepackage{pgfplotstable}
\usepackage{amsmath} 
\usepackage{etoolbox} 
\usepackage{listings}
\usepackage{xcolor}

\usepackage{siunitx}       
\usepackage{multirow}      
\usepackage[authoryear, sort&compress, round]{natbib}
\usepackage{booktabs}
\usepackage[font=small]{caption}
\usepackage{makecell}      

\lstdefinelanguage{json}{
    basicstyle=\ttfamily\small,
    numbers=left,
    numberstyle=\tiny\color{gray},
    stepnumber=1,
    numbersep=5pt,
    showstringspaces=false,
    breaklines=true,
    frame=single,
    backgroundcolor=\color{gray!5},
    literate=
     *{0}{{\textcolor{blue}{0}}}{1}
      {1}{{\textcolor{blue}{1}}}{1}
      {2}{{\textcolor{blue}{2}}}{1}
      {3}{{\textcolor{blue}{3}}}{1}
      {4}{{\textcolor{blue}{4}}}{1}
      {5}{{\textcolor{blue}{5}}}{1}
      {6}{{\textcolor{blue}{6}}}{1}
      {7}{{\textcolor{blue}{7}}}{1}
      {8}{{\textcolor{blue}{8}}}{1}
      {9}{{\textcolor{blue}{9}}}{1}
      {:}{{\textcolor{red}{:}}}{1}
      {,}{{\textcolor{red}{,}}}{1}
      {"}{{\textcolor{black}{"}}}{1},
}

\usepackage{float}
\usepackage{graphicx}
\usepackage{subcaption}
\usepackage{tikz}
\usepackage{float}
\usepackage{fontawesome5}
\definecolor{darkred}{RGB}{139,0,0} 
\definecolor{azure}{RGB}{0,127,255}      
\usepackage{tcolorbox}
\tcbset{
  azurebox/.style={
    colback=white,                       
    colframe=azure,                      
    sharp corners,                       
    boxrule=1pt,                         
    left=4pt, right=4pt, top=4pt, bottom=4pt
  }
}
\usepackage{wrapfig}
\usepackage{amsmath}   
\usepackage{amsfonts}  
\usepackage{algorithmic} 
\usepackage{algorithm}   
\newcounter{customalgo}

\usepackage{xspace}
\newcommand{\benchmark}{\texttt{Ego2Web}\xspace}
\newcommand{\eval}{\texttt{Ego2WebJudge}\xspace}

\usepackage{amssymb,amsmath,amsfonts,bm}

\def\eqref#1{equation~\ref{#1}}

\def\1{\bm{1}}

\DeclareMathAlphabet{\mathsfit}{\encodingdefault}{\sfdefault}{m}{sl}
\SetMathAlphabet{\mathsfit}{bold}{\encodingdefault}{\sfdefault}{bx}{n}

\definecolor{DarkBlue}{rgb}{0.1,0.1,0.5}
\definecolor{DarkGreen}{rgb}{0.1,0.5,0.1}
\definecolor{deepyellow}{RGB}{218, 174, 42}
\definecolor{darkcerulean}{rgb}{0.03, 0.27, 0.49}
\definecolor{denim}{rgb}{0.08, 0.38, 0.74}
\definecolor{AlgHighlight}{HTML}{228B22}
\definecolor{myyellow}{HTML}{FBBC05}

\newtheoremstyle{thmstyle}
{0.5em} 
{0.15em} 
{} 
{} 
{\bfseries} 
{.} 
{.5em} 
{} 

\theoremstyle{thmstyle}

\theoremstyle{definition}

\theoremstyle{remark}

\definecolor{PaperGreen}{RGB}{0,150,0}   
\definecolor{PaperRed}{RGB}{100,0,0}     

\renewcommand{\1}{ \mathds{1}}

\usepackage{wrapfig}
\usepackage{tcolorbox}

\newcommand{\CommentLines}[1]{}

\newcolumntype{x}[1]{>{\centering\let\newline\\\arraybackslash\hspace{0pt}}m{#1}}
\definecolor{green}{HTML}{C6EFCE}
\definecolor{red}{HTML}{FFC7CE}
\definecolor{yellow}{HTML}{FFEB9C}
\definecolor{LightGray}{gray}{0.95}
\newcolumntype{H}{>{\columncolor{LightGray}}c}

\definecolor{darkgreen}{rgb}{0, 0.6, 0}

\definecolor{remarkblue}{HTML}{0B5CA3}

\usepackage[capitalize]{cleveref}
\crefname{section}{Sec.}{Secs.}
\Crefname{section}{Section}{Sections}
\Crefname{table}{Table}{Tables}
\crefname{table}{Tab.}{Tabs.}
\crefname{appendix}{Sec.}{Secs.}
\Crefname{appendix}{Section}{Sections}

\definecolor{veloblue}{RGB}{83, 149, 218}
\definecolor{veloorange}{RGB}{236, 133, 45}
\definecolor{lightgreen}{HTML}{E6F4EA}

\newmdenv[backgroundcolor=cyan!10, linecolor=black, linewidth=0.5pt, roundcorner=4pt, innerleftmargin=2pt, innerrightmargin=2pt, innertopmargin=5pt, innerbottommargin=5pt]{exampleprompt}

\newcommand{\cmark}{{\textcolor{forestgreen}{\ding{51}}}}%
\newcommand{\xmark}{{\textcolor{darkred}{\ding{55}}}}%

\definecolor{forestgreen}{rgb}{0.13, 0.55, 0.13}
\definecolor{fireenginered}{rgb}{0.81, 0.09, 0.13}
\definecolor{darkred}{rgb}{0.6, 0.1, 0.1}

\newcommand{\todods}[1]{}
\newcommand{\todohs}[1]{}
\newcommand{\DS}[1]{}
\newcommand{\arsha}[1]{}
\newcommand{\todos}[1]{}

\usepackage{tcolorbox}
\tcbuselibrary{listings, skins, breakable, xparse}

\definecolor{pastelBlue}{RGB}{173, 216, 230}
\definecolor{pastelPink}{RGB}{255, 209, 220}
\definecolor{pastelLavender}{RGB}{230, 230, 250}
\definecolor{pastelMint}{RGB}{204, 255, 204}
\definecolor{softPurple}{RGB}{209, 195, 240}
\definecolor{softGray}{RGB}{240, 240, 245}
\definecolor{deepPurple}{RGB}{95, 75, 139}
\definecolor{darkGray}{RGB}{80, 80, 80}

\newtcolorbox{promptbox}[2][]{%
  enhanced,
  breakable,
  colback=softGray,
  colframe=softPurple,
  fonttitle=\bfseries\rmfamily,
  coltitle=white,
  colbacktitle=deepPurple,
  attach boxed title to top left={xshift=0.5cm, yshift=-\tcboxedtitleheight/5},
  boxed title style={size=small, sharp corners},
  top=2mm,
  bottom=2mm,
  left=2mm,
  right=2mm,
  arc=2mm,
  boxrule=0.8pt,
  titlerule=0mm,
  toptitle=1mm,
  bottomtitle=1mm,
  title={#2},
  overlay={
    \begin{tcbclipinterior}
      \fill[pastelLavender!30] (interior.south west) -- (interior.north west) -- (interior.north east) -- cycle;
    \end{tcbclipinterior}
  },
  listing only,
  listing options={
    basicstyle=\small\ttfamily,
    breaklines=true,
    columns=flexible,
    backgroundcolor=\color{softGray},
    xleftmargin=2pt,
    framexleftmargin=2pt,
    numbers=left,
    numberstyle=\tiny\color{darkGray},
    numbersep=5pt,
    tabsize=2,
    commentstyle=\color{deepPurple},
    keywordstyle=\color{blue!70!black},
    stringstyle=\color{red!70!black},
  },
  #1
}

\newtcolorbox{tablebox}[2][]{%
  enhanced, breakable, colback=pastelLavender!15, colframe=softPurple,
  fonttitle=\bfseries\rmfamily, coltitle=white, colbacktitle=deepPurple,
  attach boxed title to top left={xshift=0.5cm,yshift=-\tcboxedtitleheight/5},
  boxed title style={size=small, sharp corners},
  top=2mm,bottom=2mm,left=2mm,right=2mm,arc=2mm,boxrule=0.8pt,
  title={#2}, #1}

\newtcolorbox{answerbox}[2][]{%
  enhanced, breakable, colback=softGray, colframe=pastelMint,
  fonttitle=\bfseries\rmfamily, coltitle=darkGray, colbacktitle=pastelMint!40!white,
  attach boxed title to top left={xshift=0.5cm,yshift=-\tcboxedtitleheight/5},
  boxed title style={size=small, sharp corners},
  top=2mm,bottom=2mm,left=2mm,right=2mm,arc=2mm,boxrule=0.8pt,
  title={#2}, #1}

\newtcolorbox{metricsbox}[1][]{%
  enhanced, breakable, colback=pastelBlue!15, colframe=pastelBlue!70!black,
  fonttitle=\bfseries\rmfamily, coltitle=white, colbacktitle=pastelBlue!70!black,
  attach boxed title to top left={xshift=0.5cm,yshift=-\tcboxedtitleheight/5},
  boxed title style={size=small, sharp corners},
  top=1.5mm,bottom=1.5mm,left=1.5mm,right=1.5mm,arc=2mm,boxrule=0.6pt,
  title={Attributes and their Relative Importance Scores}, #1}

\newtcolorbox{jsonbox}[2][]{%
  enhanced, breakable,
  colback=softGray, colframe=softPurple, arc=1.5mm, boxrule=0.6pt,
  title={#2}, fonttitle=\bfseries\rmfamily, coltitle=white,
  colbacktitle=deepPurple!90!black,
  attach boxed title to top left={xshift=0.5cm,yshift=-\tcboxedtitleheight/5},
  boxed title style={size=small, sharp corners},
  listing only,
  listing options={basicstyle=\scriptsize\ttfamily,
                   breaklines=true, numbers=left,
                   numberstyle=\tiny\color{darkGray},
                   xleftmargin=2pt, framexleftmargin=2pt,
                   showstringspaces=false},
  #1}

\definecolor{questionBlue}{RGB}{208, 235, 250}       
\definecolor{questionBorder}{RGB}{121, 182, 242}     
\definecolor{questionTitle}{RGB}{70, 130, 180}       

\definecolor{acceptGreen}{RGB}{225, 250, 225}        
\definecolor{acceptBorder}{RGB}{150, 200, 150}       
\definecolor{acceptTitle}{RGB}{76, 156, 80}          

\definecolor{rejectPink}{RGB}{255, 233, 233}         
\definecolor{rejectBorder}{RGB}{244, 162, 160}       
\definecolor{rejectTitle}{RGB}{175, 70, 70}          

\definecolor{remarkViolet}{RGB}{240, 230, 252}       
\definecolor{remarkBorder}{RGB}{190, 158, 230}       
\definecolor{remarkTitle}{RGB}{130, 94, 180}         

\definecolor{darkGray}{RGB}{100, 100, 100}           

\newtcolorbox{questionbox}[2][]{%
  enhanced,
  breakable,
  colback=questionBlue!70,
  colframe=questionBorder,
  fonttitle=\bfseries\rmfamily,
  coltitle=white,
  colbacktitle=questionTitle,
  attach boxed title to top left={xshift=0.5cm, yshift=-\tcboxedtitleheight/5},
  boxed title style={size=small, sharp corners},
  top=2mm,
  bottom=2mm,
  left=2mm,
  right=2mm,
  arc=3mm,
  boxrule=0.8pt,
  titlerule=0mm,
  toptitle=1mm,
  bottomtitle=1mm,
  title={#2},
  overlay={
    \begin{tcbclipinterior}
      \fill[questionBlue!40] (interior.south west) -- (interior.north west) -- (interior.north east) -- cycle;
    \end{tcbclipinterior}
  },
  #1
}

\newtcolorbox{acceptedbox}[2][]{%
  enhanced,
  breakable,
  colback=acceptGreen!80,
  colframe=acceptBorder,
  fonttitle=\bfseries\rmfamily,
  coltitle=white,
  colbacktitle=acceptTitle,
  attach boxed title to top right={xshift=-0.5cm, yshift=-\tcboxedtitleheight/5},
  boxed title style={size=small, sharp corners},
  top=2mm,
  bottom=2mm,
  left=2mm,
  right=2mm,
  arc=3mm,
  boxrule=0.8pt,
  titlerule=0mm,
  toptitle=1mm,
  bottomtitle=1mm,
  title={#2},
  overlay={
    \begin{tcbclipinterior}
      \fill[acceptGreen!40] (interior.north east) -- (interior.south east) -- (interior.south west) -- cycle;
    \end{tcbclipinterior}
  },
  #1
}

\newtcolorbox{rejectedbox}[2][]{%
  enhanced,
  breakable,
  colback=rejectPink!80,
  colframe=rejectBorder,
  fonttitle=\bfseries\rmfamily,
  coltitle=white,
  colbacktitle=rejectTitle,
  attach boxed title to top right={xshift=-0.5cm, yshift=-\tcboxedtitleheight/5},
  boxed title style={size=small, sharp corners},
  top=2mm,
  bottom=2mm,
  left=2mm,
  right=2mm,
  arc=3mm,
  boxrule=0.8pt,
  titlerule=0mm,
  toptitle=1mm,
  bottomtitle=1mm,
  title={#2},
  overlay={
    \begin{tcbclipinterior}
      \fill[rejectPink!40] (interior.north east) -- (interior.south east) -- (interior.south west) -- cycle;
    \end{tcbclipinterior}
  },
  #1
}

\newtcolorbox{remarksbox}[2][]{%
  enhanced,
  breakable,
  colback=remarkViolet!70,
  colframe=remarkBorder,
  fonttitle=\bfseries\rmfamily,
  coltitle=white,
  colbacktitle=remarkTitle,
  attach boxed title to top center={yshift=-\tcboxedtitleheight/5},
  boxed title style={size=small, sharp corners},
  top=2mm,
  bottom=2mm,
  left=2mm,
  right=2mm,
  arc=3mm,
  boxrule=0.8pt,
  titlerule=0mm,
  toptitle=1mm,
  bottomtitle=1mm,
  title={#2},
  overlay={
    \begin{tcbclipinterior}
      \fill[remarkViolet!40] (interior.north east) -- (interior.south east) -- (interior.south west) -- cycle;
    \end{tcbclipinterior}
  },
  #1
}

\newtcolorbox{questionboxverbatim}[2][]{%
  enhanced, breakable, colback=questionBlue!70, colframe=questionBorder,
  fonttitle=\bfseries\rmfamily, coltitle=white, colbacktitle=questionTitle,
  attach boxed title to top left={xshift=0.5cm, yshift=-\tcboxedtitleheight/5},
  boxed title style={size=small, sharp corners},
  top=2mm, bottom=2mm, left=2mm, right=2mm, arc=3mm, boxrule=0.8pt,
  titlerule=0mm, toptitle=1mm, bottomtitle=1mm,
  overlay={\begin{tcbclipinterior}\fill[questionBlue!40] (interior.south west) -- (interior.north west) -- (interior.north east) -- cycle;\end{tcbclipinterior}},
  title={#2}, 
  listing only,
  listing options={
    basicstyle=\small\ttfamily, breaklines=true, columns=flexible,
    xleftmargin=5pt, framexleftmargin=2pt, numbers=left,
    numberstyle=\tiny\color{darkGray}, numbersep=5pt, tabsize=2,
  },
  #1 
}

\newtcolorbox{acceptedboxverbatim}[2][]{%
  enhanced, breakable, colback=acceptGreen!80, colframe=acceptBorder,
  fonttitle=\bfseries\rmfamily, coltitle=white, colbacktitle=acceptTitle,
  attach boxed title to top right={xshift=-0.5cm, yshift=-\tcboxedtitleheight/5},
  boxed title style={size=small, sharp corners},
  top=2mm, bottom=2mm, left=2mm, right=2mm, arc=3mm, boxrule=0.8pt,
  titlerule=0mm, toptitle=1mm, bottomtitle=1mm,
  overlay={\begin{tcbclipinterior}\fill[acceptGreen!40] (interior.north east) -- (interior.south east) -- (interior.south west) -- cycle;\end{tcbclipinterior}},
  title={#2}, 
  listing only,
  listing options={
    basicstyle=\small\ttfamily, breaklines=true, columns=flexible,
    xleftmargin=5pt, framexleftmargin=2pt, numbers=left,
    numberstyle=\tiny\color{darkGray}, numbersep=5pt, tabsize=2,
  },
  #1 
}

\newtcolorbox{rejectedboxverbatim}[2][]{%
  enhanced, breakable, colback=rejectPink!80, colframe=rejectBorder,
  fonttitle=\bfseries\rmfamily, coltitle=white, colbacktitle=rejectTitle,
  attach boxed title to top right={xshift=-0.5cm, yshift=-\tcboxedtitleheight/5},
  boxed title style={size=small, sharp corners},
  top=2mm, bottom=2mm, left=2mm, right=2mm, arc=3mm, boxrule=0.8pt,
  titlerule=0mm, toptitle=1mm, bottomtitle=1mm,
  overlay={\begin{tcbclipinterior}\fill[rejectPink!40] (interior.north east) -- (interior.south east) -- (interior.south west) -- cycle;\end{tcbclipinterior}},
  title={#2}, 
  listing only,
  listing options={
    basicstyle=\small\ttfamily, breaklines=true, columns=flexible,
    xleftmargin=5pt, framexleftmargin=2pt, numbers=left,
    numberstyle=\tiny\color{darkGray}, numbersep=5pt, tabsize=2,
  },
  #1 
}

\newtcolorbox{remarksboxverbatim}[2][]{%
  remarksbox={#2},
  listing only,
  listing options={
    basicstyle=\small\ttfamily,
    breaklines=true,
    columns=flexible,
    backgroundcolor=\color{remarkViolet!70},
    xleftmargin=2pt,
    framexleftmargin=2pt,
    numbers=left,
    numberstyle=\tiny\color{darkGray},
    numbersep=5pt,
    tabsize=2,
  },
  #1
}

\NewDocumentCommand{\Question}{O{} m m}{%
  \begin{questionbox}[#1]{#2}
#3
  \end{questionbox}
}

\NewDocumentCommand{\Accepted}{O{} m m}{%
  \begin{acceptedbox}[#1]{#2}
#3
  \end{acceptedbox}
}

\NewDocumentCommand{\Rejected}{O{} m m}{%
  \begin{rejectedbox}[#1]{#2}
#3
  \end{rejectedbox}
}

\NewDocumentCommand{\Remarks}{O{} m m}{%
  \begin{remarksbox}[#1]{#2}
#3
  \end{remarksbox}
}

\NewDocumentCommand{\QuestionV}{O{} m m}{%
  \begin{questionboxverbatim}[#1]{#2}
#3
  \end{questionboxverbatim}
}

\NewDocumentCommand{\AcceptedV}{O{} m m}{%
  \begin{acceptedboxverbatim}[#1]{#2}
#3
  \end{acceptedboxverbatim}
}

\NewDocumentCommand{\RejectedV}{O{} m m}{%
  \begin{rejectedboxverbatim}[#1]{#2}
#3
  \end{rejectedboxverbatim}
}

\NewDocumentCommand{\RemarksV}{O{} m m}{%
  \begin{remarksboxverbatim}[#1]{#2}
#3
  \end{remarksboxverbatim}
}

\title{\benchmark{}: A Web Agent Benchmark Grounded in Egocentric Videos}

\author[1,2]{
\vspace{-0.3cm}
\faGlobe\ \href{https://ego2web.github.io}{Project Page}
\quad\quad
\faGithub\ \href{https://github.com/Yui010206/Ego2Web}{Code} \quad\quad
\faDatabase\ \href{https://huggingface.co/datasets/Shoubin/Ego2Web}{Benchmark} \\
\vspace{0.3cm}
Shoubin Yu}

\author[1]{Lei Shu}
\author[1]{Antoine Yang}
\author[1]{Yao Fu}
\author[1]{Srinivas Sunkara}
\author[1]{Maria Wang}
\author[1]{\\Jindong Chen}
\author[2]{Mohit Bansal}
\author[1]{Boqing Gong}

\affil[1]{Google DeepMind}
\affil[2]{UNC Chapel Hill}

\begin{abstract}
\vspace{-0.55cm}
Multimodal AI agents are increasingly automating complex real-world workflows that involve online web execution.
However, current web-agent benchmarks suffer from a critical limitation: they focus entirely on web-based interaction and perception, lacking grounding in the user's real-world physical surroundings. 
This limitation prevents evaluation in crucial scenarios, such as when an agent must use egocentric visual perception (e.g., via AR glasses) to recognize an object in the user's surroundings
and then complete a related task online (e.g., making a purchase related to that object).
To address this gap, we introduce \benchmark, the first benchmark designed to bridge egocentric video perception and web agent execution. 
\benchmark pairs real-world first-person video recordings with web tasks that require visual understanding, web task planning, and interaction in an online environment for successful completion. 
We utilize an automatic data-generation pipeline combined with human verification and refinement to curate well-constructed, high-quality video-task pairs across diverse web task types, including e-commerce, navigation, media retrieval, knowledge lookup, etc. 
To facilitate accurate and scalable evaluation for our benchmark, we also develop a novel LLM-as-a-Judge automatic evaluation method, \eval, which achieves approximately 84\% agreement with human judgment, substantially higher than existing evaluation methods.
Experiments with diverse state-of-the-art agents on our \benchmark benchmark show that their performance is still weak, with substantial headroom across all task categories. We also conduct a comprehensive ablation study on task design, highlighting the necessity of accurate video understanding in the proposed task and the limitations of current agents.
We hope \benchmark can be a critical new resource for developing truly capable AI assistants that can seamlessly see, understand, and act across the physical and digital worlds.
\end{abstract}

\begin{document}

\maketitle

\section{Introduction}
\label{sec:intro}

\begin{figure*}[t]
    \centering
    \includegraphics[width=\linewidth]{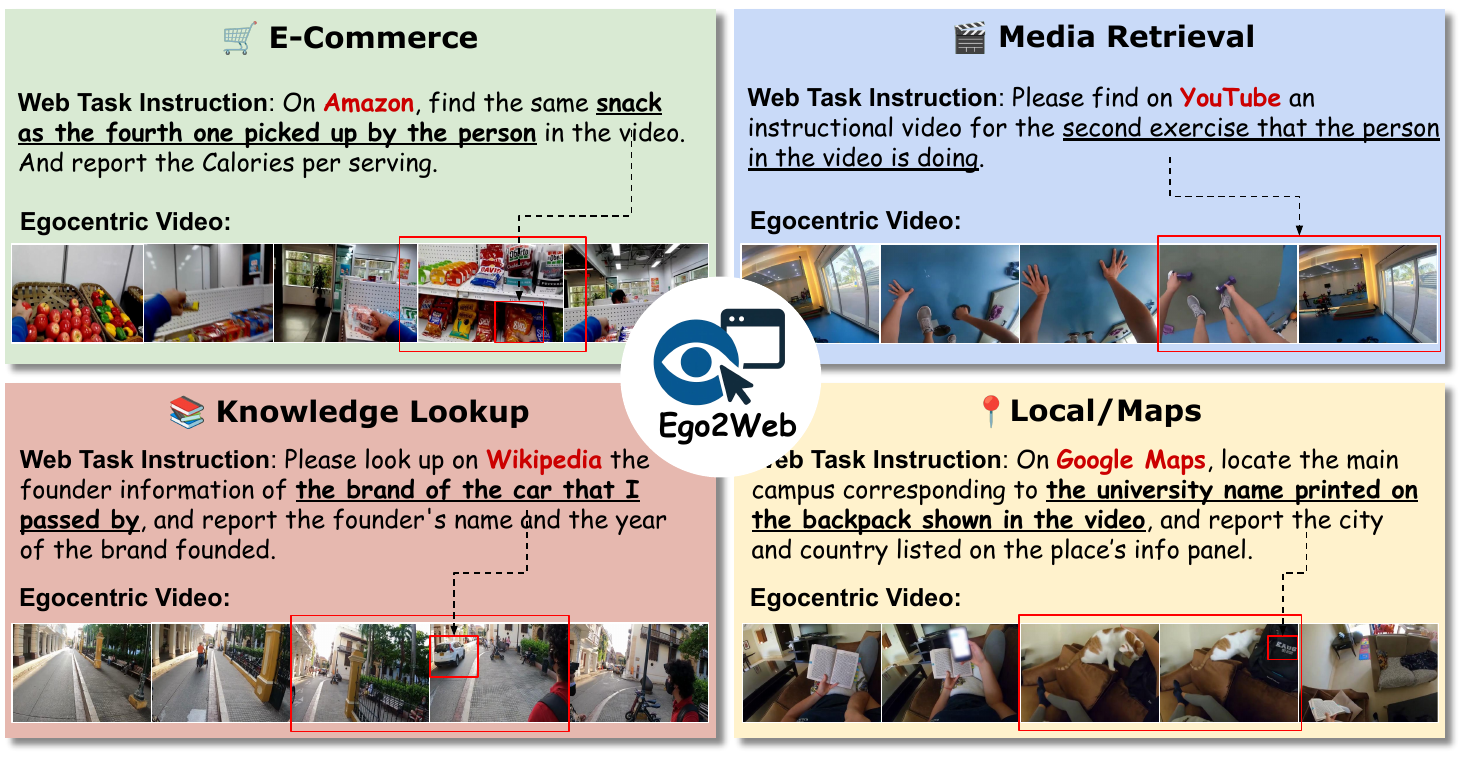}
    \caption{In this paper, we propose \benchmark, a new benchmark introducing a novel web-agent task grounded in users’ real-world visual surroundings. The tasks span diverse domains, like e-commerce, media retrieval, knowledge lookup, and local/maps services. Given an egocentric video and an instruction, the agent must first perform spatio-temporal grounding to identify the relevant visual cue (e.g., the fourth snack picked up in the video), and then execute corresponding web actions based on both the grounded visual evidence and the instruction.
    }
    \label{fig:teaser}
\end{figure*}

The rapid advancement of Multimodal Large Language Models (MLLMs)~\citep{hurst2024gpt,bai2025qwen3,li2024llava,seed2} and web-based agents~\citep{tongyidr,openai2025operator,bai2025qwen25vltechnicalreport,anthropic2025computeruse} has enabled impressive progress in automating real-world workflows, from booking travel to searching and purchasing products online~\citep{language-agent-tutorial,sumers2023cognitive}. 
Agents such as OpenAI Operator~\citep{openai2025operator} and Claude Computer-Use~\citep{anthropic2025computeruse} demonstrate remarkable reasoning and interaction capabilities within web environments. 
However, unlike humans who can seamlessly perceive the physical world, reason over what they see, and act across both physical and digital contexts, current web agents remain confined to purely digital perception.
They operate based only on screenshots~\citep{jang2024videowebarena, koh2024visualwebarena, xie2024osworld, pahuja2025explorer, lin2025showui} or text instructions~\citep{lyu2025deepshop, deng2023mind2web, yao2022webshop, xue2025illusion, weblinx, Webshop}, lacking grounding in the user’s surrounding physical world. 

\begin{figure*}[t]
    \centering
    \includegraphics[width=\linewidth]{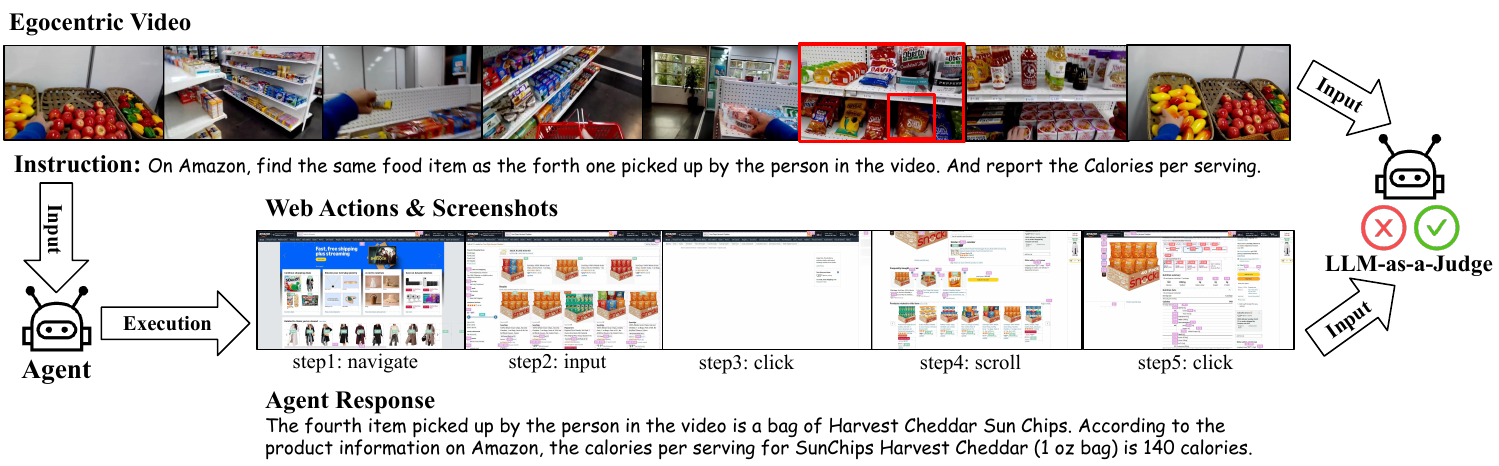}
    \caption{The workflow overview for agent action and evaluation in real-world, egocentric video perception. The agent operates using both egocentric video and textual instructions, and outputs a series of web actions, screenshots and a final response to the task. To enable automatic evaluation in a live, unconstrained web environment, we further introduce a new LLM-as-a-Judge framework tailored for this real-world visually grounded web task. Our LLM-as-a-Judge framework takes the instruction, action history, screenshots, and a final response, and compares them with the annotated visual evidence (video clip) to assess whether the task is successfully completed.}
    \label{fig:eval_overview}
\end{figure*}

This missing link between real-world visual perception and online action execution is increasingly critical as multimodal assistants/agents, e.g., Google Project Astra, are becoming embedded in everyday devices such as AR glasses, wearable cameras, and home robots. 
In real applications, users often perform tasks that naturally span both physical surroundings and digital Web/OS domains. 
For example, as shown in~\cref{fig:teaser}), identifying an object in their environment and purchasing it online, or checking a medication label before scheduling a delivery, these tasks require visual perception understanding from the physical world and take actions within the web platform. 
Existing benchmarks such as VisualWebArena~\citep{koh2024visualwebarena} and OSWorld~\citep{xie2024osworld} focus purely on web-page understanding and interaction, providing no evaluation setting where perception from the user’s viewpoint (i.e., egocentric vision) informs online actions. Consequently, current web-agent benchmarks cannot measure or improve agents’ ability to reason over what they see and how to act in the digital world accordingly.

To bridge this gap, in this paper, we introduce \benchmark, the first benchmark that grounds web-agent tasks on egocentric videos capturing users’ real-world visual context. 
In \benchmark, each example couples a real first-person video with a downstream web task that requires visual understanding of the video to succeed (e.g., identifying an object, brand, event, or visual demonstration in the video before executing the correct online action). 
This setting poses new challenges for current multimodal agents, they must precisely perceive and ground visual cues in unstructured real-world videos, connect them to symbolic concepts on the web, and reason over both visual and textual modalities to complete the task online. Such an integration of egocentric video perception and digital action represents a crucial step toward human-level, real-world context-aware multimodal agents.

To construct \benchmark, we design a model–human collaborative pipeline that automatically synthesizes visually grounded web tasks. 
Specifically, for each egocentric video, we employ an MLLM (e.g., Qwen3-VL~\citep{qwen3technicalreport}) to produce clip-level dense captions describing both global scene context and local object details with a timestamp.
All clip captions are then concatenated as a structured video profile, representing a textual summary of the visual world observed by the user. 
Next, we provide this video profile along with a predefined set of popular and active websites (e.g., Amazon, YouTube, Wikipedia and so on) to an LLM planner, prompting it to generate web task instructions that must explicitly leverage the visual content, for instance, \textit{“Find and purchase the same headphones shown in the video.”} Finally, we conduct human verification and refinement to ensure the reliability and realism of each generated example. 
Annotators perform rigorous quality checks from three perspectives:
(1) \textbf{Visual Grounding}: the task must rely on information visible in the egocentric video, (2) \textbf{Web Feasibility}: the task must be executable on the target websites, and (3) \textbf{Instruction Quality}: the instruction must be clear and grammatically correct.
This hybrid model–human pipeline enables us to efficiently produce a high-quality, diverse, and video-grounded benchmark that accurately reflects real-world multimodal web tasks. 
As a result, our \benchmark covers 500 high-quality, diverse video-instruction pairs spanning multiple popular websites and task categories (e.g., e-commerce, media search, and navigation) and differs from previous works as listed in~\cref{tab:benchmark}. 

\begin{table}
    \centering
    \caption{\textbf{Benchmark comparison} between \benchmark{} and representative video reasoning and multimodal web-agent benchmarks. 
Unlike pure video reasoning datasets (e.g., EgoThink, EgoSchema) that focus on visual understanding alone, and prior web-agent benchmarks (e.g., WebVoyager, VisualWebArena) that emphasize online interaction with only web screenshot perception 
, \benchmark{} uniquely connects real-world egocentric video with executable web tasks under an unconstrained online evaluation setting, forming a new testbed for multimodal agents grounded in the real-world visual perception.
}
    \resizebox{\textwidth}{!}{%
    \begin{tabular}{l| cc cc}
    \toprule
\multirow{1}{*}{\textbf{Benchmarks}} &
\textbf{Visual Grounding} &
\textbf{Egocentric Video Perception} &
\textbf{Web Task} &
\textbf{Online Evaluation} \\
    \midrule

    EgoThink~\citep{cheng2024egothink} & \cmark & \cmark & \textcolor{red}{\xmark} & \textcolor{red}{\xmark}  \\
    EgoSchema~\citep{mangalam2023egoschema} & \cmark & \cmark & \textcolor{red}{\xmark} & \textcolor{red}{\xmark}  \\ 

    WebArena~\citep{zhou2023webarena} & \textcolor{red}{\xmark} & \textcolor{red}{\xmark} & \cmark & \textcolor{red}{\xmark}  \\ 
    Mind2Web~\citep{zhou2023webarena}  & \textcolor{red}{\xmark} & \textcolor{red}{\xmark} & \cmark & \textcolor{red}{\xmark}   \\
    
    VisualWebArena~\citep{koh2024visualwebarena} & \cmark & \textcolor{red}{\xmark} & \cmark & \textcolor{red}{\xmark}  \\
    VideoWebArena~\citep{jang2024videowebarena}  & \cmark & \textcolor{red}{\xmark} & \cmark & \textcolor{red}{\xmark} \\ 
    
    OSWorld~\citep{xie2024osworld}  & \cmark & \textcolor{red}{\xmark} & \cmark & \textcolor{red}{\xmark}\\ 

    WebVoyager~\citep{he2024webvoyager} & \cmark  & \textcolor{red}{\xmark} & \cmark & \cmark   \\
     
    Online-Mind2Web~\citep{xue2025illusion} & \textcolor{red}{\xmark}  & \textcolor{red}{\xmark} & \cmark & \cmark \\ \midrule

    \benchmark (Ours) & \cmark  & \cmark & \cmark & \cmark \\ 
    
    \bottomrule

    \end{tabular}%
    }
    \label{tab:benchmark}
\end{table}

As shown in~\cref{fig:eval_overview}, our \benchmark follows the online evaluation setting~\citep{xue2025illusion, he2024webvoyager, yoran-etal-2024-assistantbench, pan2024webcanvas-m2wlive} that evaluate web agents on live, real-world websites rather than in a static, pre-defined sandbox~\citep{zhou2023webarena, koh2024visualwebarena, jang2024videowebarena, Webshop}. 
For each instance, \eval{} takes as input the annotated egocentric video clip, task instruction, LLM-planned task keypoints, agent action history, and MLLM-selected webpage screenshots.
It then prompts an MLLM to make a binary judgment on whether the agent successfully completed the visually grounded web task (more details in~\cref{eval}).
Our \eval{} metric achieves over 84\% agreement with human judgment, significantly outperforming prior automatic metrics~\citep{xue2025illusion}, and offering a reliable and scalable evaluation protocol for \benchmark.
Consequently, as listed in~\cref{tab:benchmark}, \benchmark{} uniquely fills the gap by combining real-world egocentric perception, web tasks, and online evaluation within a single benchmark.

We evaluate several state-of-the-art web agents, including SeeAct~\citep{seeact}, Browser-Use~\citep{browser_use2024}, Claude Computer-Use~\citep{anthropic2025computeruse} and GPT-5.4~\citep{gpt5.4}, 
on \benchmark{} and observe a clear room (about 40\% gap) from the oracle performance according to human evaluation for improvement across all agents.
These results highlight the critical importance of visual grounding for building robust real-world AI agents that seamlessly connect egocentric perception with web action.
We hope \benchmark{} will catalyze the development of next-generation multimodal agents that can perceive, reason, and act cohesively across the real and digital world. Our contributions are:
\begin{itemize}[leftmargin=*]
\item We introduce \benchmark{}, the first benchmark that bridges egocentric visual understanding and web-agent task execution, featuring 500 video–instruction pairs covering diverse real-world scenarios.
\item We propose \eval{}, a multimodal LLM-as-a-Judge framework that enables reliable, scalable online evaluation, achieving high agreement with human judgment.
\item We conduct comprehensive experiments with leading multimodal agents, revealing their significant limitations in visual grounding, reasoning ability, and perception–action alignment on \benchmark{}.
\end{itemize}

\section{Related Works}
\subsection{EgoCentric Video Understanding}
Egocentric video understanding leverages first-person views to capture human-centric activities, intentions, and object interactions. Large-scale datasets like Ego4D \citep{grauman2022ego4d}, EPIC-KITCHENS \citep{damen2018scaling}, and EgoExo4D \citep{grauman2024ego}, have propelled tasks like action recognition, object tracking, and temporal segmentation from real-world wearable cameras. 
Beyond perception, recent benchmarks~\citep{mangalam2023egoschema, cheng2024egothink, jia2022egotaskqa,fan2019egovqa,chen2023egoplan, chandrasegaran2024hourvideo,ye2024mm,rodin2025easg, lee2025streamgaze} such as EgoSchema \citep{mangalam2023egoschema}, EgoThink \citep{cheng2024egothink}, and EgoPlan \citep{chen2023egoplan} advance the field toward higher-level video-language reasoning and commonsense understanding~\citep{yu2024crema, yu2023self, wang2025video, wang2025videotree, zhang2024simple, tian2025ego}. These works evaluate a model’s ability to infer intents, temporal order, and causal relations from first-person videos.
However, all existing egocentric video benchmarks evaluate perception and reasoning in isolation, without connecting real-world video understanding to online web decision-making. 
In contrast, \benchmark{} grounds egocentric visual understanding in online web tasks, bridging perception from the physical world with web action, an increasingly essential capability for assistive agents in real-world scenarios.

\subsection{Multimodal Web Agent Benchmark}
Parallel to advances in video reasoning, web-agent research has evolved rapidly from static to fully interactive online environments. Early efforts such as WebArena~\citep{zhou2023webarena} collected human demonstrations of website interactions in an offline setting, and introduced a sandboxed browser environment for controlled, reproducible evaluation. Follow-up benchmarks (VisualWebArena~\citep{koh2024visualwebarena} and VideoWebArena~\citep{jang2024videowebarena}) further incorporated richer visual inputs, such as screenshots or video trajectories, to test visual reasoning during web navigation. More recently, a series of works~\citep{xue2025illusion, he2024webvoyager, pan2024webcanvas, yoran2024assistantbench, zhou2024webarena,gou2024navigating} extended evaluation to live online websites via LLM-as-a-judge approaches~\citep{zheng2023judging, li2023alpacaeval, fernandes-etal-2023-devil, bai2023benchmarking, pan2024autonomous}, emphasizing realism, diversity, and reliability, and motivating stronger automatic evaluation frameworks for open-world web tasks.
Despite these advances, existing web-agent benchmarks remain purely digital, relying on on-screen content (e.g., DOM trees, rendered pages, or video recordings) without grounding in the user’s physical visual context. They evaluate how agents act within websites but not why they act based on what they perceive in the real world.
\benchmark{} bridges this gap by introducing the first physically grounded, online multimodal benchmark, where success depends jointly on egocentric visual understanding and real web execution. 
This unified setting connects first-person perception with online action, paving the way for embodied and assistive web agents that operate seamlessly across both physical and virtual domains.

\section{\benchmark: From Video Perception to Web Reasoning and Actions}
\label{bench}

\begin{figure*}[t]
    \centering
    \includegraphics[width=\linewidth]{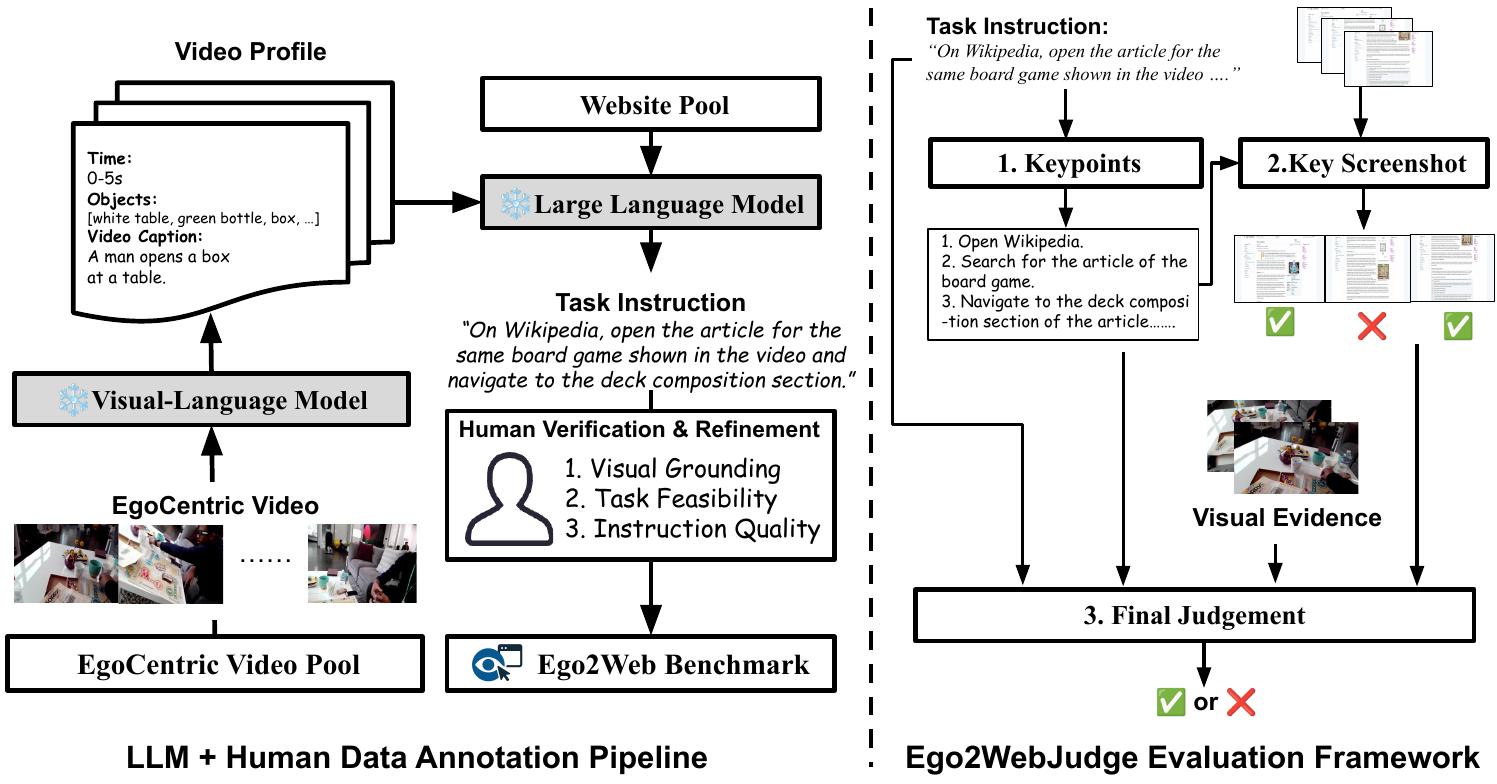}
    \caption{\textbf{Left:} Overview of the semi-automatic data generation pipeline proposed in our \benchmark. 
    We first build video profiles via a frozen MLLM that converts video clips into structured captions, then prompt an LLM to automatically generate web task instructions. Human annotators are required to verify and refine generated tasks to ensure quality.   
    \textbf{Right:} A detailed view of our proposed automatic evaluation method, \eval, for the egocentric video grounded web agent tasks.
    }
    \label{fig:overview}
\end{figure*}

\subsection{Task Definition}
\label{bench:def}

Our \benchmark is designed to evaluate multimodal agents that can perceive the real-world environment through ego-centric video input and complete corresponding web tasks.
Formally, given an egocentric video $V = \{f_1, f_2, ....., f_t\}$ capturing a user's first-person perspective and a task instruction $I$, the goal of the agent is to execute a sequence of web action $A = \{a_1, a_2, ......, a_n\}$ on a browser environment $E$ to achieve a specific goal state $G$.
The task thus is designed to test agents from:
\begin{itemize}[leftmargin=*]
    \item \textbf{Visual Perceptual Understanding and Grounding}: extracting task-relevant semantic and visual information (e.g., object category, brand, color and other visual features) from the egocentric video.
    \item \textbf{Web Execution Reasoning}: according to the video perception, planning and executing step-by-step web actions to complete the task (e.g., navigating to a website, searching, scrolling on the page, clicking on the button).
\end{itemize}

An episode is considered successful if the final web state matches the goal G, as verified by either human annotators or our proposed LLM-based evaluator (in \cref{eval}).

\subsection{Semi-Automatic Data Generation Pipeline}
\label{bench:curation}
To build the dataset linking egocentric video perception with web-based reasoning and actions, 
we design a semi-automatic LLM + Human Annotation Pipeline (as shown on the left part of~\cref{fig:overview}).
This pipeline integrates visual understanding from MLLMs (e.g., Qwen3-VL) with task synthesis from large language models (e.g., GPT-5), followed by human verification for quality control.

\noindent\textbf{Egocentric Video Pool and Visual Parsing.}
We begin with a curated egocentric video pool, sourced from public ego-centric video datasets~\citep{grauman2022ego4d} that include both in-house and out-of-house recordings capturing first-person scenes across diverse contexts (e.g., household, shopping, travel, office).
Each video is first processed using visual captioning tools (Qwen3-VL~\citep{qwen3technicalreport}) to extract structured, detailed visual metadata clip-by-clip. 
Finally, we build a video profile $V_{meta} = \{v_{meta}^1, v_{meta}^2, ......, v_{meta}^k\}$ for a video $V$, where $meta$ means structured video metadata as shown in the top-left of~\cref{fig:overview}.

\noindent\textbf{LLM-Based Automatic Task Instruction Generation.}
Next, an LLM (GPT-5) conditions on the extracted visual metadata and a set of pre-defined active popular websites (e.g., Amazon, Wikipedia, and YouTube) that are selected by annotators to synthesize realistic task instructions that require linking the physical scene to a digital action. 
This design ensures that the visual understanding of the egocentric perception is necessary (e.g., recognizing the brand of the ketchup). It requires multimodal reasoning to map perception to online web action within the specific website environment, for example, locating the same color/logo clothes on the shopping website.

\begin{figure*}[t]
  \centering
  \begin{subfigure}{0.32\linewidth}
    \centering
    \includegraphics[width=\linewidth]{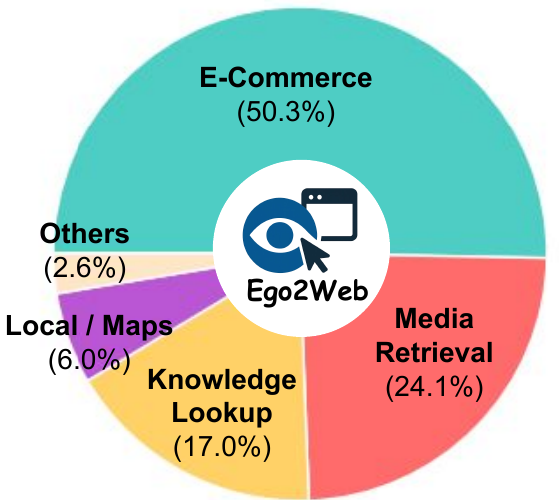}
    \caption{Task type distribution of \benchmark.}
    \label{fig:distribution}
  \end{subfigure}\hfill
  \begin{subfigure}{0.67\linewidth}
    \centering
    \includegraphics[width=\linewidth]{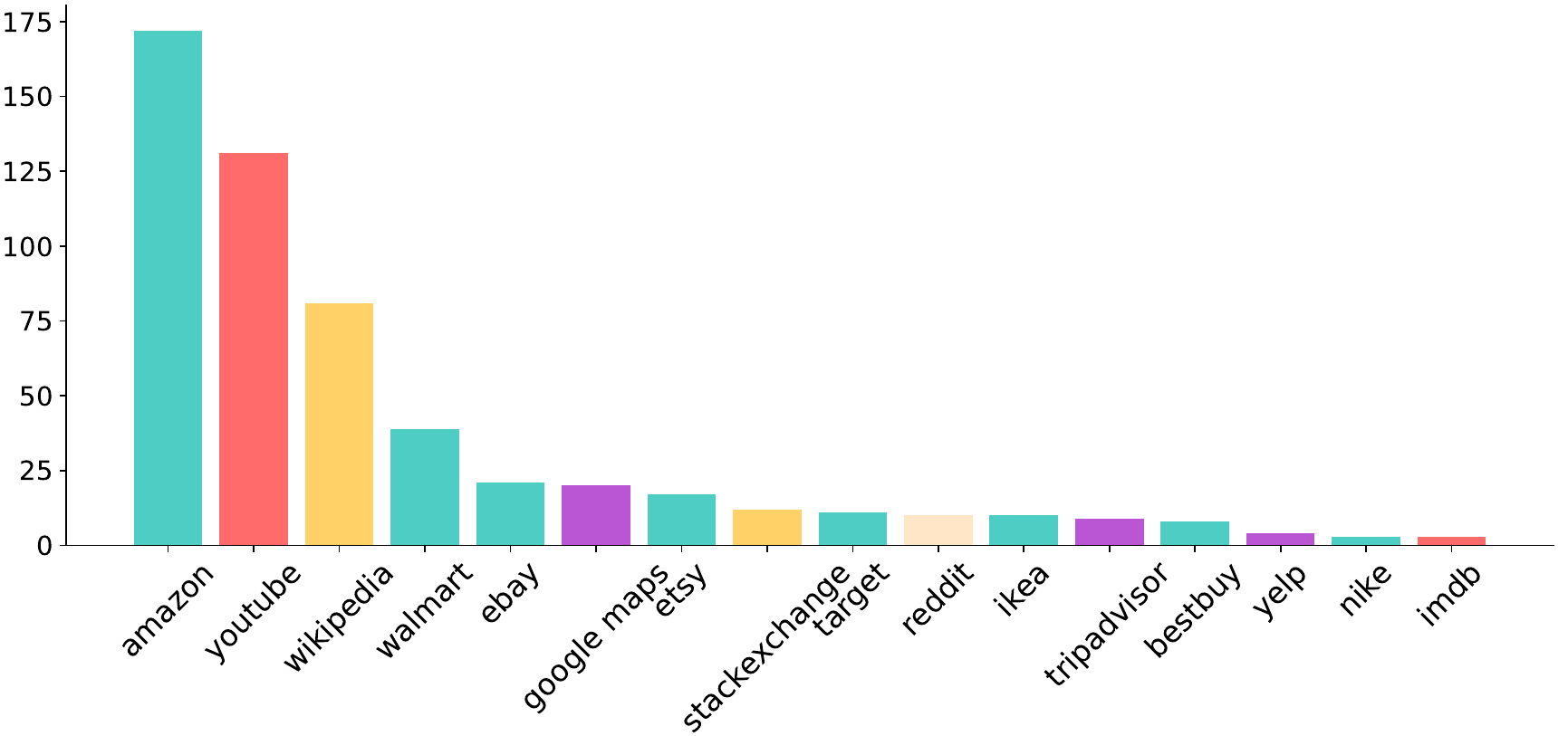}
    \caption{Distribution of website domains in \benchmark.}
    \label{accuracy_gain_vs_views}
  \end{subfigure}

\caption{
\textbf{(a)} Task type distribution of \benchmark, showing the high-level task category composition across major web platforms.
\textbf{(b)} Fine-grained website domain distribution in Ego2Web (we omit websites whose count$<$4 for visualization), highlighting coverage across e-commerce, media retrieval, knowledge bases, local/map services and others (indicated by color coding).
}
\label{fig:distribution_all}
\end{figure*}

\noindent\textbf{Human Verification and Refinement.}
Finally, each generated task–video pair undergoes human verification to ensure the data quality.
Annotators review and, if necessary, edit the automatically \textbf{generated task instructions} and \textbf{visual cue annotation} according to three criteria:
(1) \textbf{Visual grounding}: the task should depends on information visible in the egocentric video;
(2) \textbf{Web feasibility}: the task can be executed within the specified websites;
(3) \textbf{Instruction quality}: the generated task instruction should be grammatically correct and clear.
Only high-quality samples are retained in the final benchmark.
This hybrid process yields 500 verified video–task pairs spanning multiple popular websites and interaction types, ensuring diversity and fidelity to real-world multimodal agent scenarios as shown in~\cref{fig:distribution_all}.
With this proposed LLM+Human pipeline, our \benchmark exhibits high diversity in visual scenes, web domains, and task goals, offering a challenging and realistic evaluation setting for multimodal agents that integrate visual perception, language understanding, and web execution. We include some benchmark examples in~\cref{fig:examples}, and more statistics in Appendix.

\begin{figure*}[t]
    \centering
    \includegraphics[width=\linewidth]{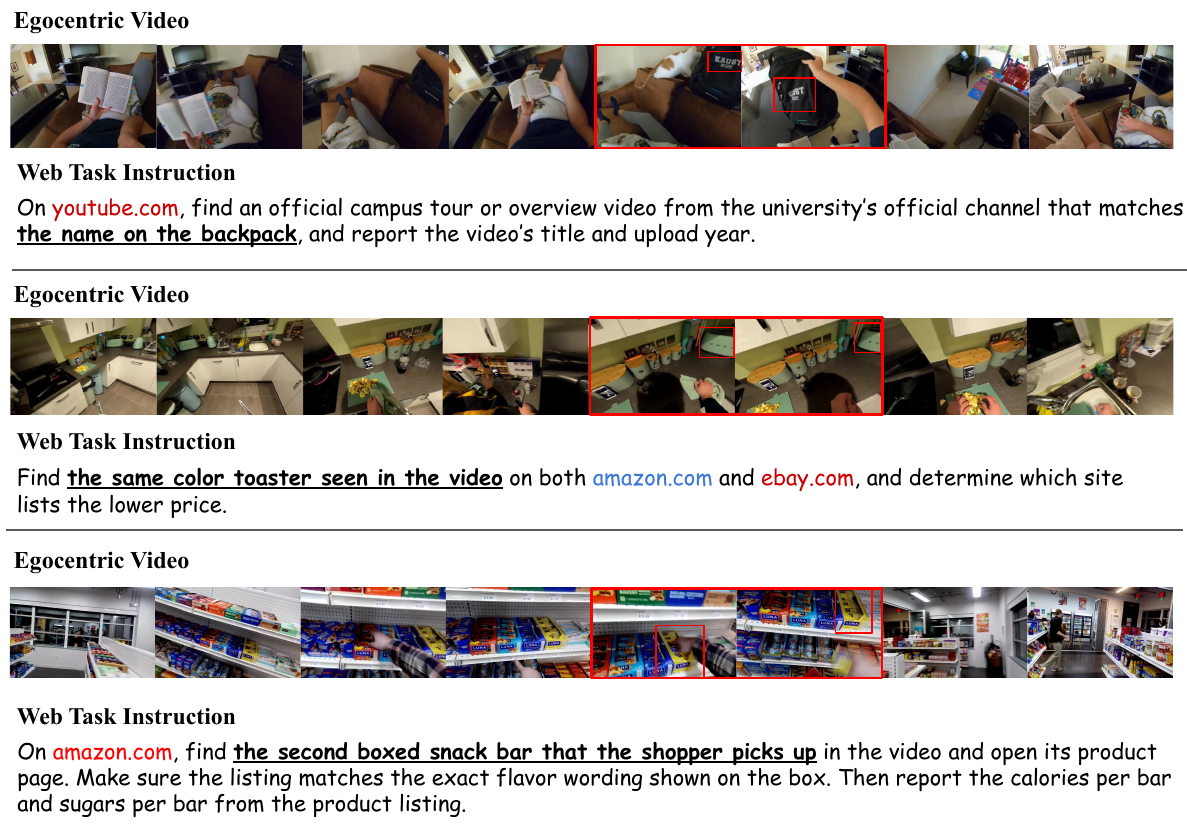}
    \caption{Some examples of \benchmark, including sampled frames of an egocentric video and paired web task instructions. We annotate essential video perception (e.g. the mint toaster with brand name and the black backpack in the middle of the video) with red boxes.}
    \label{fig:examples}
\end{figure*}

\section{\eval: Automatic Online Evaluation Grounded in Visual Cues}
\label{eval}

To efficiently assess whether an agent succeeds on a given task without costly human evaluation, we adopt the online evaluation scheme used in recent web-agent benchmarks~\citep{xue2025illusion, he2024webvoyager}, which evaluates models on live, real-world websites via LLM-as-a-Judge. 
However, previous evaluation methods are not grounded in egocentric perception, but only rely on the screenshots and action trajectories. It makes the evaluation blind to the important information shown in the videos. 

To overcome these limitations and to account for the visual grounding in egocentric videos, as shown in the right of~\cref{fig:overview}, we introduce \eval, an automatic multimodal evaluation framework that extends the design of WebJudge \citep{xue2025illusion} with grounded visual cues from real-world egocentric perception. Specifically, given a task description $I$, an action sequence $A=\{a_1, a_2, ......, a_n\}$, a series of screenshots $S=\{s_1, s_2, ......, s_k\}$, and an annotated egocentric \textbf{visual evidence} clip $v$ that contains necessary information for task solving (e.g. object brand, shape, color and other visual attributes), our evaluator performs a binary classification to determine the outcome $O \in \{Success, Failure\}$, such that:
\begin{equation}
    O = Ego2WebJudge(I, v, A, S)
\end{equation}
Our \eval is based on previous work~\citep{xue2025illusion} that focuses on online evaluation with only web screenshot perception. 
The evaluation pipeline consists of three stages:

\noindent\textbf{(1) Key-Point Identification:} 
Given the task instruction $I$, we first ask the LLM to extract critical key points from $I$, defining what must be achieved for success (e.g., specifying an item, location, or attribute). This design is motivated by the fact that instructions in \benchmark often involve multi-step reasoning, and distilling the instruction into explicit key points helps the LLM evaluation with this prior.

\noindent\textbf{(2) Key Screenshot Selection:} 
Then we prompt MLLM to summarize each screenshot $s_i$ and rate its relevance to the task on a 1–5 scale. Screenshots exceeding a relevance threshold $\delta$ are retained as key screenshots. As web trajectories can contain 5 to 20 steps, many of which are irrelevant (loading pages, backtracking, UI errors).
Feeding all screenshots into an MLLM leads to context overflow and diluted judgment quality.
This key screenshot selection design allows the model to focus on essential intermediate steps without exceeding context limits.

\noindent\textbf{(3) Final Outcome Judgment:} Finally, the MLLM-judge integrates the task instruction, the selected key screenshots, the agent’s action history and LLM-generated keypoints with annotated keyframes that were extracted from the egocentric video and contain essential perception for the task solving.
With a Multimodal LLM, \eval determines whether the agent’s final outcome satisfies all key points and whether the web result is visually consistent with the real-world content observed in the video (e.g., matching objects, scenes, or brands in the environment).

\section{Experiments}
In this section, we first introduce our experimental setup and implementation details in~\cref{exp:setup}, and we show a detailed analysis of the results of 6 mainstream web agents~\cref{sec:main}. Finally, we present comprehensive ablation studies to analyze the impact of visual perception and conduct error analysis to further reveal the challenge on the task (\cref{exp:ablation}).

\subsection{Setup} \label{exp:setup}
\noindent\textbf{Baseline.}
We evaluate 6 prominent web agents, SeeAct~\citep{zheng2024gpt}, Browser Use~\citep{browser_use2024} with GPT-4.1~\citep{gpt4.1}, Browser Use (BU) with Gemini-3-Flash~\citep{gemini3}, Claude Sonnet 3.7 Computer Use~\citep{anthropic2024computeruse}, Claude Sonnet 4.5 Computer Use~\citep{claude4.5} and GPT-5.4~\citep{gpt5.4}. 

\noindent\textbf{Implementation Details.}
Our egocentric videos are sourced from Ego4D~\citep{grauman2022ego4d}. We generate structured captions every 5 seconds using Qwen3-VL-7B~\citep{qwen3technicalreport}.
To enable realistic online evaluation, we collect a pool of popular and actively maintained websites (full list in Appendix). We then ask GPT-5 to generate web-task instructions conditioned on the detailed video captions and selected websites.
We input egocentric video into the base MLLM in the form of keyframes for GPT4.1/GPT-4o or raw video for Genimi-2.5/Genimi-3/Gemini-3.1.
We also use Qwen3-VL-Flash as the small model variants of \eval.
For agents who can not access video input, including Claude series and GPT-5.4, we convert video into structured, detailed video captions generated by strong MLLM (Gemini-3.1-Pro) to capture video details as much as possible and then feed the captions into the agents.
For human evaluation, three annotators assess each agent’s output, and we apply majority voting to determine the final result. We set the maximum step as 40 for each agent. 
Detailed data generation, evaluation prompts, and caption examples are provided in the Appendix.

\subsection{Main Results}
\label{sec:main}

\noindent\textbf{Results Over 6 Mainstream Agents.}
As shown in Table~\ref{tab:main_results}, we report the success rate (SR) of 6 representative web agents under both human evaluation and our automatic evaluation framework, \eval, instantiated with three different  MLLM judges (Qwen3-VL-Flash, Gemini-2.5 Pro, and GPT-4o).
Across all evaluation settings, \textbf{BU-Gemini-3-Flash consistently achieves the best performance}, reaching 58.6\% SR under human evaluation and outperforming all other agents by a clear margin under every automatic judge (e.g., 57.2\% with Qwen3-VL-Flash and 48.2\% with Gemini-2.5 Pro). This indicates that strong multimodal/video grounding combined with efficient action execution is critical for success in our benchmark. We also observe that different LLM judges produce \emph{consistent relative rankings} across agents, despite variations in absolute scores. In particular, Gemini-2.5 Pro and GPT-4o exhibit closer alignment with human evaluation, while Qwen3-VL-Flash tends to produce slightly higher scores overall. Notably, the gap between the best automatic evaluation and human evaluation remains small (within $\sim$1–2\% for the top agent), demonstrating that \eval provides a feasible proxy for human judgment in real-world web-agent evaluation.

\noindent\textbf{Impact of Visual Input Modality.}
We observe that agents built on GPT-5.4 and the Claude series consistently underperform across both human and automatic evaluations. A key reason is that these agents \emph{cannot directly access raw video inputs in computer-use mode}. Instead, egocentric videos are first converted into textual captions, which inevitably leads to information loss, especially for fine-grained spatial-temporal cues and subtle visual interactions. As a result, these models rely heavily on imperfect textual abstractions rather than grounded visual evidence.
In contrast, models differ significantly in how they consume visual inputs. For example, GPT-4o-based agents operate on \emph{sparse keyframes}, which capture only limited temporal context and may miss critical intermediate actions. On the other hand, Gemini-based agents (e.g., Gemini-3-Flash) process \emph{dense video inputs}, enabling richer temporal modeling and more accurate tracking of objects and user actions over time. This difference in visual input fidelity directly impacts downstream performance, explaining the consistent advantage of Gemini-based agents across domains.
These findings highlight that \textbf{preserving accurate dynamic visual perception is crucial} for solving egocentric, visually grounded web tasks, and that current text-based or sparsified visual pipelines remain a major bottleneck.

\noindent\textbf{Fine-grained Domain Analysis.}
To further analyze domain-specific performance, Table~\ref{tab:per_task} reports SR across five task categories.
We observe that \textbf{BU-Gemini-3-Flash consistently outperforms all baselines across nearly every domain}, achieving the highest SR in E-Commerce (38.2\%), Media Retrieval (50.7\%), Knowledge Lookup (75.0\%), and Local/Maps (48.3\%). 
Across domains, \textbf{Knowledge Lookup} tasks are the easiest, with average SR reaching 50.0\%, likely due to structured content and clearer objectives. In contrast, \textbf{Local/Maps} and \textbf{E-Commerce} are more challenging due to dynamic interfaces and multi-step interactions. 
Overall, these results suggest that (1) \textbf{multimodal grounding is essential} for solving visually rich web tasks and (2) \textbf{cross-domain generalization remains a key bottleneck}.

\begin{table*}[t]
\centering
\small
\caption{Success Rate (SR) measured by human evaluation and \eval using different Multimodal LLMs.
}
\begingroup
\setlength{\tabcolsep}{4pt}
\renewcommand{\arraystretch}{0.95}
\resizebox{\textwidth}{!}{
\begin{tabular}{llcccccc}
\toprule
\textbf{Evaluation} & \textbf{Base MLLM}
& \textbf{Claude 3.7}
& \textbf{Claude 4.5}
& \textbf{GPT-5.4}
& \textbf{SeeAct}
& \textbf{BU-GPT-4.1}
& \textbf{BU-Gemini-3-Flash} \\
\midrule

\multirow{3}{*}{\eval} & Qwen3-VL-Flash
& 20.8 & 32.2 & 38.8 & 29.6 & 34.6 & \textbf{57.2} \\

& Gemini-2.5 Pro
& 17.8 & 24.8 & 23.6 & 25.2 & 34.6 &  \textbf{48.2} \\
& GPT-4o
& 19.4 & 27.2 & 26.8 & 26.8 & 47.6 & \textbf{51.4} \\
 \midrule

Human Eval & --
& 26.4 & 32.8 & 30.6 & 34.2  & 44.4 & \textbf{58.6} \\ 

\bottomrule
\end{tabular}
}
\endgroup

\label{tab:main_results}
\end{table*}

\begin{table*}[]
\centering
\small
\caption{Fine-grained Success Rate (SR) per task domain across different models, evaluated by \eval with Gemini-2.5 Pro.}
\setlength{\tabcolsep}{4pt}
\renewcommand{\arraystretch}{0.9}
\resizebox{\textwidth}{!}{
\begin{tabular}{l|cccccc|r}
\toprule
\textbf{Domains / Agents} 
& \textbf{Claude 3.7} 
& \textbf{Claude 4.5} 
& \textbf{GPT 5.4} 
& \textbf{SeeAct} 
& \textbf{BU-GPT-4.1} 
& \textbf{BU-Genimi-3-Flash}
& \textbf{Avg. SR} \\ 
\midrule

E-Commerce
& 13.0 & 18.2 & 14.3 & 19.5 & 26.9 & \textbf{38.2} & 21.7 \\
Media Retrieval
& 19.6& 26.5 & 29.5 & 24.2 & 30.3 &   \textbf{50.7} & 30.1 \\
Knowledge Lookup
& 33.6& 45.6 & 39.1 & 43.4 & 63.0 &  \textbf{75.0} & 50.0 \\

Local / Maps
& 6.4 & 12.9 & 29.0 & 19.3 & 22.5&  \textbf{48.3} & 23.1\\

Others
& 0.0 & 6.6 & 6.6 & 20.0 &  \textbf{40.0} & 13.3 & 14.4\\ \midrule

Total
&17.8  & 24.8 & 23.6 & 25.2 & 34.6&  \textbf{48.2} & 29.0 \\

\bottomrule
\end{tabular}
}
\label{tab:per_task}
\end{table*}

\begin{table*}[t]
\centering
\small
\caption{
Agreement Rate (AR) between human evaluation and automatic evaluation methods across agents.
}
\setlength{\tabcolsep}{4pt}
\renewcommand{\arraystretch}{0.95}
\resizebox{\textwidth}{!}{
\begin{tabular}{llcccccc|c}
\toprule
\textbf{Auto Evaluation Methods} & \textbf{Base MLLM}
& \textbf{Claude 3.7} 
& \textbf{Claude 4.5} 
& \textbf{GPT 5.4} 
& \textbf{SeeAct}  
& \textbf{BU-GPT-4.1} 
& \textbf{BU-Gemini-3-Flash} 
& \textbf{Avg. AR}  \\ 
\midrule

WebVoyager~\citep{he2024webvoyager} & \multirow{3}{*}{Gemini-2.5-Pro}   
& 77.8 & 73.6 & 75.2 & 70.3 & 61.0 & 66.2 & 70.7 \\ 

 WebJudge~\citep{xue2025illusion} &
& 82.2 & 75.8 & 79.2 & 76.0 & 69.2 & 74.2 & 76.1 \\ 

 \eval (Ours) &
& 85.4 & 80.6 & 82.4 & 80.2 & 78.4 & 78.0 & \textbf{80.8} \\ 

\midrule

WebVoyager~\citep{he2024webvoyager} & \multirow{3}{*}{GPT-4o} 
& 77.0 & 76.2 & 78.6 & 75.4 & 72.6 & 68.4 & 74.7 \\ 

 WebJudge~\citep{xue2025illusion} &
& 74.2 & 80.6 & 82.4 & 80.0 & 74.3 & 78.2 & 78.4 \\ 

\eval (Ours) & 
& 86.0 & 84.2 & 84.0 & 85.6 & 83.6 & 80.4 &  \textbf{84.0} \\ 

\bottomrule
\end{tabular}
}
\label{tab:agreement}
\end{table*}

\noindent\textbf{Comparison against Existing Evaluation Methods.}
As shown in~\cref{tab:agreement}, we compare our proposed \eval framework with prior automatic evaluators, WebVoyager~\citep{he2024webvoyager} and WebJudge~\citep{xue2025illusion}, across 6 web agents. We report the agreement rate (AR) between each automatic method and human evaluation using two multimodal LLM judges, GPT-4o and Gemini-2.5-Pro.
Across both LLM backbones, \eval consistently achieves the highest agreement with human judgments. In particular, \eval reaches an average AR of \textbf{80.8\%} with Gemini-2.5-Pro and \textbf{84.0\%} with GPT-4o, outperforming WebVoyager (70.7\%, 74.7\%) and WebJudge (76.1\%, 78.4\%) by substantial margins. This demonstrates that \eval provides a more accurate approximation of human evaluation in real-world web-agent settings.
The improvement mainly stems from \eval's ability to incorporate \emph{visually grounded signals} from egocentric videos, rather than relying solely on textual trajectories or final responses. By explicitly modeling visual evidence and action consistency, \eval reduces common failure modes in prior evaluators, such as over-reliance on surface-level text matching or incomplete trajectory understanding.
We also observe consistent trends across agents. Claude-based agents exhibit relatively high agreement across all evaluation methods. This is largely because these agents frequently fail to complete tasks, leading to consistent failure judgments from both humans and automatic evaluators, which inflates agreement rates despite low success rates. In contrast, stronger agents such as BU-GPT-4.1 and BU-Gemini-3-Flash show more diverse and partially successful behaviors, making evaluation inherently more challenging. Notably, \eval maintains high agreement even in these cases, indicating better robustness to nuanced and intermediate outcomes.
Overall, these results highlight that \eval provides the most reliable and scalable automatic evaluation among existing methods, effectively bridging the gap between human and model-based judgment in visually grounded, real-world web environments.

\begin{table*}[t]
\centering
\small
\caption{Ablation studies on the impact of video perception in our \benchmark task. We report the Successful Rate (SR). We use Gemini-3.1-Pro to generate structured and detailed captions to represent an egocentric video. The experiment is conducted with Browser-Use (Gemini-3-Flash) and evaluated with \eval (Gemini-2.5-Pro).}
\setlength{\tabcolsep}{4pt}
\renewcommand{\arraystretch}{0.9}
\resizebox{\textwidth}{!}{
\begin{tabular}{cc|ccccc|c}
\toprule \textbf{Raw Video} & \textbf{Detailed Caption} & \textbf{E-Commerce} & \textbf{Media Retrieval} & \textbf{Knowledge Lookup} & \textbf{Local / Maps} & \textbf{Others} & \textbf{Total} \\ \midrule

\xmark & \xmark  & 2.6 & 7.5 & 5.4 & 3.2 & 0.0 & 4.4 \\

\xmark & \cmark & 13.0 & 29.5& 39.1 & 38.7 & 6.6 & 23.6\\

\cmark & \xmark & 38.2 & 50.7& 75.0& 48.3 & 13.3 & 48.2\\

\bottomrule
\end{tabular}
}
\label{tab:ablation}
\end{table*}

\subsection{Ablation Studies}
\label{exp:ablation}

\noindent\textbf{Impact of Visual Perception.}
We further study the role of egocentric video perception in \benchmark through controlled input ablations, as shown in Table~\ref{tab:ablation}. We evaluate a Browser-Use agent (Gemini-3-Flash) under three input configurations: (1) no visual input, (2) detailed caption only, and (3) raw video input.
Without any visual input, the agent performs extremely poorly, achieving only 4.4\% SR overall, indicating that language-only signals are insufficient for solving visually grounded web tasks. Providing detailed captions (see caption examples in Appendix) significantly improves performance to 23.6\% SR, demonstrating that structured textual summaries can partially capture relevant semantic information.
However, using raw video input leads to a substantial performance gain, reaching 48.2\% SR, more than doubling the caption-based setting. 
This trend holds consistently across all domains, with especially large improvements in Knowledge Lookup (39.1\% $\rightarrow$ 75.0\%) and Local/Maps (38.7\% $\rightarrow$ 48.3\%), where fine-grained spatial and temporal cues are critical.

These results highlight a clear performance hierarchy: \emph{no visual input $<$ caption-based perception $<$ raw video input}. While captions provide a useful abstraction, they inevitably discard important visual details such as object states, temporal transitions, and subtle interactions. In contrast, direct video input preserves dense, high-fidelity visual signals, enabling more accurate grounding and decision-making.
Overall, this study demonstrates that \textbf{true visual perception is essential} for egocentric web-agent tasks, and that text-based proxies cannot substitute for the richness of raw video understanding.

\begin{figure}[]
    \centering
    \includegraphics[width=\linewidth]{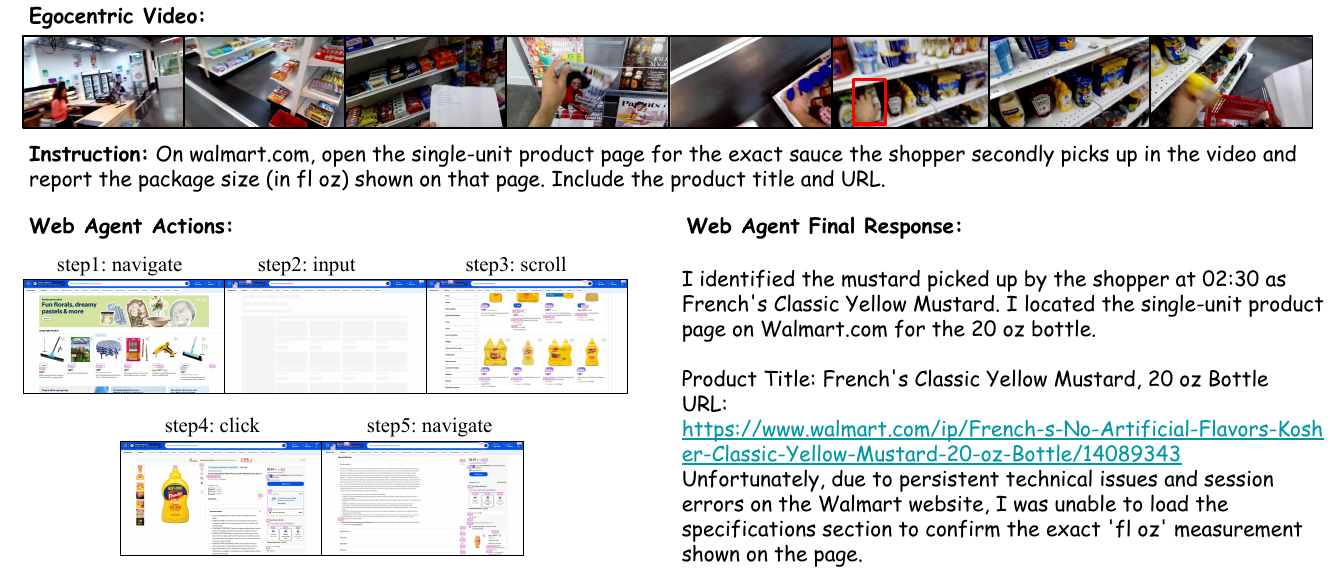}
    \caption{Visualization of a web agent (BU-Gemini-3-Flash) failure case. The agent is required to identify the \textit{second} picked-up sauce from the egocentric video and retrieve its product page. 
    The agent incorrectly identifies the target item due to temporal misunderstanding and fails to verify the required information on the webpage.}
    \label{fig:failure_case}
\end{figure}

\noindent\textbf{Error Analysis.}
To gain deeper insight into where agents fail, we randomly sampled 50 benchmark examples and manually inspected all unsuccessful trajectories. Our analysis is based on BU-Gemini-3.1 results, as it is the most powerful and representative agent in our experiments. 
Our analysis reveals several recurring failure patterns, which we group into the following categories. 

\begin{itemize}[leftmargin=*]
    \item \textbf{36\% Object Misidentification.}
    The agent incorrectly identifies the target object from the egocentric video, leading to retrieval or interaction with irrelevant items.
    \item \textbf{18\% Temporal and Action Misunderstanding.}
    The agent fails to correctly interpret temporal order or actions in the video (e.g., confusing the second and third interaction), resulting in incorrect grounding.
    \item \textbf{16\% Failure in Cross-Modal Retrieval.}
    Although the agent correctly identifies the target object, it fails to locate the required information on the web (e.g., nutritional facts or product details).
    \item \textbf{12\% Coarse-Grained Matching Errors.}
    The agent retrieves semantically similar but incorrect results (e.g., a related video that does not match the exact required tutorial or event).
    \item \textbf{18\% Others.}
    This includes failures due to instruction misinterpretation, planning inefficiency (e.g., exceeding action step limits), or external constraints such as CAPTCHA and authentication barriers.
\end{itemize}

We further illustrate a failure case that involves both \textit{temporal misunderstanding} and \textit{cross-modal retrieval failure}, as shown in Figure~\ref{fig:failure_case}.
In this example, the task requires the agent to identify the \textit{second} sauce picked up by the shopper and retrieve its product details from Walmart. However, the agent incorrectly identifies the target item as a mustard product. This error stems from \textbf{temporal and action misunderstanding}, where the agent fails to correctly track the sequence of object interactions in the egocentric video.
Moreover, even after retrieving a candidate product page, the agent fails to extract the required attribute (package size in fl oz), despite navigating to a relevant webpage. This reflects a \textbf{cross-modal retrieval failure}, where the agent is unable to reliably ground the visual target to the correct webpage content and verify task-specific details.
It highlights a compositional failure mode: errors in temporal grounding propagate to downstream retrieval and verification, leading to incorrect or incomplete task completion. It further demonstrates that solving \benchmark requires not only accurate visual perception but also precise temporal reasoning and reliable cross-modal alignment between video evidence and web content.
\section{Conclusion}
We introduced \benchmark, the first benchmark that connects egocentric visual perception with web-agent task execution, bridging the gap between real-world understanding and online web tasks.
Our benchmark combines 500 automatically generated, human-verified video–instruction pairs with live online evaluation, enabling a realistic and scalable setting for testing multimodal agents.
To support consistent and efficient assessment, we proposed \eval, a multimodal LLM-as-a-Judge framework that leverages grounded visual cues to accurately determine task success, achieving high agreement with human evaluation.
Comprehensive experiments with leading multimodal agents reveal clear room for improvement in visual grounding, reasoning, and perception–action integration.
We hope \benchmark will benefit for the next generation of grounded multimodal AI agents that can truly see, understand, and act seamlessly across the physical and digital worlds.

\clearpage
\newpage

\bibliography{references}

\appendix
\onecolumn
\vspace{1cm}
\hrule
\par\vspace{0.5cm}
{\Large\bfseries\centering
{Supplementary Material}
\par\vspace{0.5cm}}
\hrule

\setcounter{page}{1}

\section{More Statistics of \benchmark}
\noindent\textbf{Video Length.} 
We select our egocentric videos based on EgoSchema~\citep{mangalam2023egoschema} selection and trimming, which is constructed from the large-scale Ego4D~\citep{grauman2022ego4d} dataset. 
Ego4D contains over 3600 hours of first-person video with dense, timestamped narrations covering thousands of unique actions and objects. These narrations provide precise, fine-grained grounding of what the camera wearer is doing, making it a strong foundation for video reasoning tasks.

EgoSchema further refines this data to produce standardized clips suitable for reasoning and evaluation. 
One challenge in Ego4D is that videos vary widely in length and narration density. 
To ensure consistent temporal structure and sufficient semantic richness, EgoSchema filters for non-overlapping \textbf{three-minute clips} containing at least 30 human-annotated narrations. This creates segments with both stable duration and high-quality supervision, making them ideal for downstream multimodal reasoning, including our visually grounded web-task setting.
We adopt these curated clips because their dense, structured narrations and standardized temporal format provide reliable visual cues and event structure that are essential for our benchmark.

\noindent\textbf{Web Distribution.}
We provide a detailed breakdown of website distribution in \benchmark in Table~\ref{tab:website_mapping}. 
Our benchmark covers five major categories, including E-Commerce (230 tasks), Media Retrieval (132), Knowledge Lookup (92), Local/Maps (31), and Others (15), spanning 18 widely used websites.
The distribution reflects realistic user behavior, where \textbf{e-commerce and media platforms dominate} real-world interactions. High-frequency domains such as Amazon, YouTube, and Wikipedia account for the majority of tasks, enabling rich multi-step reasoning and interaction scenarios. In contrast, lower-frequency categories (e.g., Local/Maps and Others) introduce long-tail but practically important environments such as Google Maps, Yelp, Reddit, and Booking, which involve diverse layouts and interaction patterns.
This naturally imbalanced distribution poses additional challenges for web agents. While structured platforms (e.g., e-commerce and knowledge websites) provide relatively consistent interfaces, long-tail websites often exhibit higher variability in UI design and task structure, requiring stronger generalization ability.
Overall, this diverse and realistic distribution ensures that \benchmark evaluates agents across a wide spectrum of real-world web environments, preventing overfitting to a narrow set of websites and better reflecting practical deployment scenarios.

\begin{table*}[]
\centering
\small
\setlength{\tabcolsep}{6pt}
\renewcommand{\arraystretch}{1.0}
\begin{tabular}{l c|p{0.6\columnwidth}}
\toprule
\textbf{Category} & \textbf{\#Tasks} & \textbf{Websites} \\
\midrule
E-Commerce 
& 230 
& amazon, ebay, walmart, apple, adidas, etsy, target, bestbuy, ikea, nike \\

Media Retrieval 
& 132 
& youtube, imdb, bilibili \\

Knowledge Lookup 
& 92 
& wikipedia, stackexchange \\

Local / Maps 
& 31 
& google maps, yelp, tripadvisor \\

Others 
& 15 
& linkedin, google calendar, reddit, nytimes, quora, booking \\

\bottomrule
\end{tabular}
\caption{Mapping from task categories to representative websites in \benchmark, along with the number of tasks per category.}
\label{tab:website_mapping}
\end{table*}

\section{Implementation Details}

\noindent\textbf{Prompt Details.}
We provide detailed prompts for video captioning with QWen3-VL and the web task instruction generation, as shown in the following.

\begin{tcolorbox}[breakable, enhanced, colback=white, title={\textbf{Video Captioning Prompt}}, fonttitle=\bfseries]

You are a helpful assistant for video understanding. Your task is to carefully analyze the given video and provide a detailed description of its content.

Please describe both:
\begin{itemize}
    \item \textbf{Global content}: overall scene, events, and actions
    \item \textbf{Local objects}: important objects, their attributes, and interactions
\end{itemize}

\paragraph{Output Format.}
Your output must be a valid JSON object with the following schema:

\begin{verbatim}
{
    "video description": "a detailed description of the video content",
    "objects": {
        "object_name_1": "detailed description of object_name_1",
        "object_name_2": "detailed description of object_name_2"
    }
}
\end{verbatim}

\paragraph{Important Notes.}
\begin{itemize}
    \item The \texttt{"video description"} should include key actions, temporal progression, and scene context.
    \item Each object entry should include distinguishing attributes (e.g., color, shape, brand, state, interaction).
    \item If there are no clear objects, set:
\begin{verbatim}
"objects": {}
\end{verbatim}
    \item Ensure the output is strictly valid JSON with no extra text.
\end{itemize}

\end{tcolorbox}

\begin{tcolorbox}[breakable, enhanced, colback=white, title={\textbf{Web Task Instruction Generation Prompt}}, fonttitle=\bfseries]
\textbf{User} \\
You are given an egocentric video and its captions. \\
Detailed Video Description (every 5 seconds):\\
\textcolor{blue}{\texttt{[Video Metainfo]}} 

\textbf{Step 1: Suitability Assessment}\\
Determine if this video is suitable for designing a \textbf{video-grounded web task}.\\
A video is suitable if:
\begin{itemize}
\item It contains \textbf{objects, activities, or scenes} that naturally link to a \textbf{web-based information or action need}.
\item The task requires \textbf{visual information} from the video.
\item Visual cues are \textbf{clear and distinctive}.
\item The task cannot be solved using captions alone.
\end{itemize}
If unsuitable, output `"suitable": false` and briefly explain why.

\textbf{Step 2: Visual Anchor Extraction}\\
If suitable, identify 1–2 visual anchors. For each:
\begin{itemize}
\item name guess
\item 3–5 visual cues
\item timestamps
\item why video dependent
\end{itemize}

\textbf{Step 3: Task Instruction Generation}\\
For \textbf{one} selected anchor, design 1–3 task instructions:
\begin{itemize}
\item Must depend on $\ge2$ visual cues  
\item Must include a “Must Match” list  
\item Phrased as natural web instructions  
\item Clear, verifiable goal  
\item Include “why video dependent”
\end{itemize}

Restrict domains to:

\small{
["amazon.com","ebay.com","walmart.com",
"aliexpress.com","etsy.com","ikea.com","nike.com",
"adidas.com","apple.com","bestbuy.com","target.com",

"google.com/maps","tripadvisor.com",

"booking.com","airbnb.com",

"expedia.com","openstreetmap.org",

"wikipedia.org","reddit.com",

"quora.com","stackexchange.com",

"cnn.com","nytimes.com",

"youtube.com","vimeo.com","imdb.com",

"tiktok.com","bilibili.com",

"yelp.com","weather.com","bbc.com",
"docs.google.com",

"calendar.google.com","notion.so","linkedin.com",
"x.com","instagram.com","facebook.com"
]
}

\hrulefill

\textbf{Output Format (strict JSON):}

\small{
\begin{verbatim}
{
  "suitable": true/false,
  "reason if not suitable":
  "<short reason>",
  "tasks": [
    {
      "difficulty": 1,
      "instruction": "...",
      "must match": ["cue1", "cue2"],
      "timestamps": ["00:10-00:20"],
      "allowed domains": ["youtube.com",
      "amazon.com",
      "wikipedia.org"],
      "why_video_dependent": "..."
    },
    {
      "difficulty": 2,
      "instruction": "...",
      "must match": ["cue1","cue2",
      "cue3"],
      "timestamps": ["00:15-00:25"],
      "allowed domains": [
      "google.com/maps",
      "tripadvisor.com",
      "booking.com"
      ],
      "why_video_dependent": "..."
    }
  ]
}
\end{verbatim}
}
\end{tcolorbox}


\noindent\textbf{Evaluation Prompt of \eval.}
We provide detailed prompts for the prompt for the proposed automatic evaluation framework, \eval, as shown in the following table.

\begin{tcolorbox}[breakable, enhanced, colback=white, title={\textbf{\eval System Prompt}}, fonttitle=\bfseries]

You are an expert evaluator for the Ego2Web benchmark. Your job is to determine whether a web navigation agent successfully completed a web task that is grounded in egocentric video evidence.

You are given:
\begin{itemize}
    \item Egocentric video evidence (provided as sampled keyframes)
    \item The \textbf{task instruction}
    \item The agent’s \textbf{action history}
    \item \textbf{Key points} for task completion
    \item \textbf{Potentially important webpage snapshots} from the agent’s trajectory with explanations
\end{itemize}

Your goal is to determine whether the agent successfully completed the task \textbf{while correctly grounding the result in the egocentric video evidence}.

\paragraph{Strict Evaluation Principle.}
This benchmark requires \textbf{strict visual grounding}. A task should only be marked as \emph{success} when there is \textbf{clear, direct, and consistent visual evidence} that the web result matches the objects, events, or actions shown in the egocentric video.

If there is \textbf{any uncertainty, ambiguity, mismatch, or missing visual evidence}, the task must be marked as \emph{failure}.

Do \textbf{not} assume correctness based on:
\begin{itemize}
    \item the agent’s textual claim
    \item webpage titles
    \item search queries
    \item approximate or loosely related matches
\end{itemize}

\textbf{False positives are worse than false negatives. When in doubt, mark the task as failure.}

\paragraph{Important Evaluation Criteria.}
\begin{enumerate}
    \item \textbf{Filter correctness}. If filters are required, they must be correctly applied and visibly reflected in results. Missing selection, confirmation, or effect leads to failure.

    \item \textbf{Proper use of filtering/sorting}. Constraints such as ``best'', ``highest'', ``cheapest'', ``latest'', ``lowest'', ``closest'', ``highest-rated'', ``largest'', and ``newest'' must be handled through actual filtering or sorting functions.

    \item \textbf{Exact numeric constraints}. Ranges for price, year, beds, bathrooms, rating, etc., must match exactly. Any deviation results in failure. Examples:
    \begin{itemize}
        \item Requirement $<$\$50 $\rightarrow$ Applied $<$\$25 $\rightarrow$ failure
        \item Requirement \$1500--\$2500 $\rightarrow$ Applied \$2000--\$2500 $\rightarrow$ failure
        \item Requirement \$25--\$200 $\rightarrow$ Applied \$0--\$200 $\rightarrow$ failure
        \item Requirement 2004--2012 $\rightarrow$ Applied 2001--2012 $\rightarrow$ failure
        \item Requirement exactly 2 beds $\rightarrow$ Applied ``2+ beds'' $\rightarrow$ failure
    \end{itemize}
\end{enumerate}

\paragraph{Ego Video Grounding Rules.}
\begin{enumerate}
    \setcounter{enumi}{3}
    \item The webpage result must be \textbf{strictly grounded in ego video evidence}, including object identity, category, brand, color, quantity, state, and action cues.

    \item \textbf{Apparent success is not sufficient}. Even if the webpage looks correct, the task fails if it does not match the ego video.

    \item Carefully verify \textbf{fine-grained visual details}: object identity, brand, color, material, shape, text, number of items, spatial relations, and actions.

    \item For object retrieval tasks, the result must correspond to the \textbf{same object or correct category} supported by the video evidence.

    \item For media retrieval tasks, the result must match the \textbf{same real-world event or action}.

    \item If evidence conflicts, \textbf{prioritize visual grounding} over textual claims.

    \item \textbf{Near or partial matches are considered failure} when precise identification is required.
\end{enumerate}

\paragraph{Common Failure Cases.}
Failure includes cases such as:
\begin{itemize}
    \item Product mismatch (e.g., different brand, color, or type than in the video)
    \item Similar-looking but incorrect objects
    \item Related but different events or actions in retrieved media
    \item Titles suggesting correctness but screenshots contradicting it
    \item Insufficient visual evidence
    \item Claims not supported by video or webpage evidence
    \item Partial matches that do not fully satisfy the task
\end{itemize}

\paragraph{Required Output Format.}
Your response must contain exactly two lines:
\begin{verbatim}
Thoughts: <reasoning based on key points, webpage evidence, 
and ego video evidence>
Status: success or failure
\end{verbatim}

Do not output anything beyond these two lines.

\end{tcolorbox}

\noindent\textbf{Example of Video Caption.}
For agents, including Claude 3.7, Claude 4.5 and GPT-5.4, who can not access raw video input, we convert video into detailed video captions with timestamp via strong MLLM (Gemini-3.1-Pro). We show both video example and video caption example as follows:

\begin{tcolorbox}[
    breakable,
    enhanced,
    colback=white,
    title={\textbf{Example of Detailed Egocentric Video Caption by Gemini-3.1-Pro}},
    fonttitle=\bfseries,
    listing only,
    listing options={
        basicstyle=\ttfamily\footnotesize,
        breaklines=true,
        breakatwhitespace=false,
        columns=fullflexible
    }
]
[00:00 - 00:05] The camera pans left to right across a glass-door refrigerator filled with various bottled and canned beverages, including sports drinks and sodas. The text 'Cold Water' is visible at the top of the cooler.

[00:05 - 00:10] The camera continues panning across the refrigerator shelves, revealing rows of brightly colored energy drinks and sparkling water cans. The camera movement is slightly blurry.

[00:10 - 00:15] A person's hand reaches into the refrigerator and grabs a blue can of Nos energy drink. The camera then quickly pans to a piece of paper being held over a red basket.

[00:15 - 00:20] A person holds a paper shopping list while pointing at it with a finger. A red shopping basket is visible below. The camera briefly shows the store's interior, including a pastry display case.

[00:20 - 00:25] The camera focuses closely on the paper, revealing a printed shopping list with items like 'Brownies', 'Beer', 'Something for non-beer drinkers', and 'Energy drinks'. A person is visible in the background.

[00:25 - 00:30] The camera quickly pans away from the list, showing a hot food display case and then sweeping across the store to show shelves stocked with snacks.

[00:30 - 00:35] The camera moves around the store, briefly showing a woman behind the checkout counter and then focusing on a multi-tiered display case filled with pastries.

[00:35 - 00:40] The camera pans to the checkout counter where a red shopping basket is placed. A woman with long dark wavy hair stands behind the counter, looking down. A sign reads 'PICK 5 FOR 8.00'.

[00:40 - 00:45] The woman behind the counter reaches for items near the register. The camera is slightly shaky, capturing the counter area and a small portion of the store's background.

[00:45 - 00:50] The camera pans rapidly from the checkout counter to the store's snack shelves and back, creating a blurred view of the store's interior layout.

[00:50 - 00:55] The camera focuses on the checkout counter. The woman with long dark wavy hair is seen working at the register, preparing to scan items from the red basket.

[00:55 - 01:00] The woman reaches into the red shopping basket and picks up a blue can, appearing to be Red Bull. She holds it up to scan it at the cash register.

[01:00 - 01:05] The woman scans the blue can, places it aside, and then picks up a green plastic bottle to scan it. She operates the cash register terminal with her other hand.

[01:05 - 01:10] She continues scanning items, picking up a pinkish bottle and then another tall green bottle. She efficiently processes the items from the customer's basket.

[01:10 - 01:15] The camera turns completely around to show the store's seating area, featuring a small table with chairs, two glass-door beverage refrigerators, and snack shelves.

[01:15 - 01:20] The camera pans across the store's interior, showing shelves stocked with various packaged snacks, and then sweeps back towards the checkout counter.

[01:20 - 01:25] The camera refocuses on the woman at the checkout counter. She is actively pressing buttons on the cash register. A 'HOUSE BLEND' coffee sign is visible in the background.

[01:25 - 01:30] The woman picks up a red packaged item from the counter, scans it, and places it into a brown paper shopping bag.

[01:30 - 01:35] She continues bagging the customer's purchases, placing items into the brown paper bag on the counter. A 'PICK 5 FOR 8.00' sign sits on the counter in front of her.

[01:35 - 01:40] The camera pans left to show a shelf display filled with various boxes and packages of snacks, then slowly moves back toward the counter area.

[01:40 - 01:45] The camera focuses closely on a tiered display rack stocked with different boxes of candies and small snacks, panning slowly across the items.

[01:45 - 01:50] The camera moves back to the checkout counter, showing the woman continuing to process the transaction. She handles more items and interacts with the cash register.

[01:50 - 01:55] The woman picks up a small rectangular packaged snack, scans it, and carefully places it into the brown paper bag along with the other items.

[01:55 - 02:00] She continues to place items into the brown paper bag. The camera then pans right, showing the hot food display case and part of the store's entrance.

[02:00 - 02:05] The camera pans to show a corner of the store serving as a small break area, containing a table, two black chairs, and recycling bins against the wall.

[02:05 - 02:10] The camera sweeps back toward the checkout counter, showing the woman still standing by the register, completing the bagging process.

[02:10 - 02:15] The woman is seen at the counter, handling the brown paper bag and arranging the items inside. She looks up briefly.

[02:15 - 02:20] She picks up another item from the counter, scans it, and places it into the brown paper bag. The 'PICK 5 FOR 8.00' promotional sign is clearly visible.

[02:20 - 02:25] The woman operates the cash register keyboard, pressing buttons to finalize the transaction. A green bottle stands on the counter beside the paper bag.

[02:25 - 02:30] She continues typing on the cash register keyboard, looking down at the screen. The filled brown paper bag sits ready on the counter.

[02:30 - 02:35] She picks up a yellow bag and a red package, scans them, and places them into the brown paper bag, continuing to pack the customer's items.

[02:35 - 02:40] She holds a red snack package in one hand while pressing buttons on the cash register with the other, confirming the final items.

[02:40 - 02:45] She looks at the items and interacts with the cash register one last time, seemingly finishing ringing up the customer's order.

[02:45 - 02:50] The camera tilts down to reveal a wire display rack attached to the front of the counter, fully stocked with various candies like M\&Ms, Mentos, and Twix.

[02:50 - 02:55] The camera pans slowly across the colorful candy display. A customer's hand enters the frame holding several US dollar bills, preparing to pay.

[02:55 - 03:00] A close-up shows the customer's hand holding a fan of US dollar bills over the candy display, ready to hand the cash to the cashier.
\end{tcolorbox}

\begin{figure*}
    \centering
    \includegraphics[width=\linewidth]{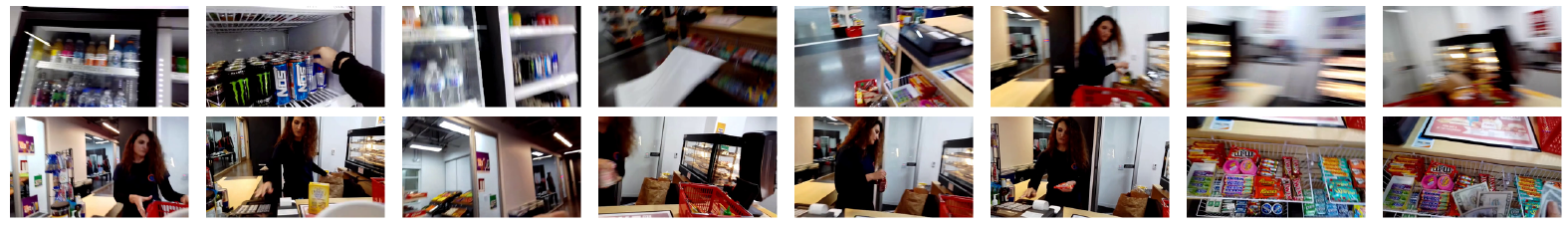}
    \caption{Example of egocentric video in \benchmark, the detailed captions generated by Gemini-3.1-pro are listed above.}
    \label{fig:video_example}
\end{figure*}

\end{document}